\documentclass[11pt]{article}

\usepackage{amsmath,amsbsy,amsfonts,amssymb,amsthm,dsfont,fullpage,units}

\usepackage[utf8]{inputenc} 
\usepackage[T1]{fontenc}    
\usepackage{hyperref}       
\usepackage{url}            
\usepackage{booktabs}       
\usepackage{amsfonts}       
\usepackage{nicefrac}       
\usepackage{microtype}      
\usepackage{natbib}
\usepackage{amsmath,amssymb}
\usepackage{graphicx}
\usepackage{multirow}
\usepackage{subfigure}
\usepackage{bbm}
\usepackage{xcolor}
\usepackage{algorithm, algorithmic}
\graphicspath{{figures/}}
\newcommand{\R}{\mathbb{R}}
\newcommand{\N}{\mathcal{N}}
\newcommand{\loss}{\mathcal{L}}
\newcommand{\hz}{\hat{z}}
\newcommand{\deps}{0.0051}
\DeclareMathOperator*{\argmin}{arg\,min}
\DeclareMathOperator*{\argmax}{arg\,max}  

\newcommand{\E}{\mathbb{E}}

\newcommand{\gen}{G}
\newcommand{\class}{C_\theta}
\newcommand{\eps}{\varepsilon}
\newcommand{\norm}[2]{{\|{#1}-{#2}\|}_2}

\title{The Robust Manifold Defense: \\  Adversarial Training using Generative Models}

\author{Ajil Jalal\footnote{Equal Contribution} \\ \texttt{ajiljalal@utexas.edu} \\ UT Austin  \and Andrew Ilyas\footnotemark[1] \\ \texttt{ailyas@mit.edu} \\ MIT EECS \and 
Constantinos Daskalakis \\ \texttt{costis@csail.mit.edu} \\ MIT EECS \and Alexandros G. Dimakis \\
\texttt{dimakis@austin.utexas.edu} \\ UT Austin}
\begin{document}

\maketitle

\begin{abstract}

We propose a new type of attack for finding adversarial examples for image classifiers. Our method exploits spanners, i.e.~deep neural networks whose input space is low-dimensional and whose output range approximates the set of images of interest. Spanners may be generators of GANs or decoders of VAEs. The key idea in our attack is to search over latent code pairs to find ones that generate nearby images with different classifier outputs. We argue that our attack is stronger than searching over perturbations of real images. Moreover, we show that our stronger attack can be used to reduce the accuracy of Defense-GAN to 3\%, resolving an open problem from the well-known paper by Athalye et al. We combine our attack with normal adversarial training to obtain the most robust known MNIST classifier, significantly improving the state of the art against PGD attacks. Our formulation involves solving a min-max problem, where the min player sets the parameters of the classifier and the max player is running our attack, and is thus searching for adversarial examples in the {\em low-dimensional} input space of the spanner.\footnote{All code and models are available at \url{https://github.com/ajiljalal/manifold-defense.git}
}









\end{abstract}

\section{Introduction}
Deep neural network (DNN) classifiers are  demonstrating excellent performance in various computer vision tasks. These models work well on benign inputs but recent work has shown that it is possible to make very small changes to an input image and drastically fool state-of-the-art models~\citep{szegedy2013intriguing,goodfellow2014explaining}.
These \textit{adversarial examples} are barely perceivable by humans, can be targeted to create desired labels
even with black-box access to  classifiers, and can be implemented as objects in the physical world
~\citep{papernot2016practical,kurakin2016adversarial,athalye2017synthesizing}.

In this work, we propose a novel optimization perspective towards obtaining stronger attacks and stronger defenses. Our starting point is the assumption that we have access to a ``spanner'' for the type of data that we are considering. A spanner is a DNN $G: \mathbb{R}^k \rightarrow \mathbb{R}^n$, whose input space is low-dimensional and whose output range $\{G(z)\}_{z \in \mathbb{R}^k}$ is a good approximation to the dataset of interest. A spanner could be the generator of a GAN or the decoder of an auto-encoder, which are trained on the same dataset that the classifier being defended or attacked was trained on. We are interested in answering the following question:

\begin{center}
\textit{Can we use a spanner to obtain improved defenses or attacks of DNN based classifiers?}
\end{center}

To answer this question, we first propose a preliminary defense approach, called Invert-and-Classify (INC), which simply uses the spanner as a ``denoiser.'' Given an image $x$ we find its projection $G(z^*)$ to the range of the spanner,\footnote{$G(z^*)$ is the projection of an image $x$ to the range of the spanner $G$ iff $z^*=\argmin_z\|x-G(z)\|$, i.e., $G(z^*)$ is the image in the range of the spanner that is closest to $x$.} and then apply the classifier $C$, which is being defended, on the projection $G(z^*)$, i.e.~we output $C(G(z^*))$. We show that this attack is not robust by proposing our ``overpowered latent-space attack," which successfully circumvents this defense. 

We note that a one-sided variant of our overpowered attack has been used by~\citet{athalye2018obfuscated} to partially circumvent the DefenseGAN defense~\citep{samangouei2018defensegan}, which amounts precisely to the INC defense described above, when the spanner is instantiated as a generator from a GAN. 

Our overpowered attack is formulated as a min-max optimization problem, where the adversary has more power than usual: rather than perturbing a single image $x$, we attempt to identify a pair of latent codes $z,z'$ such that 
the  images generated from these codes are similar, i.e.~$||G(z)-G(z')||$ is small, but the classifier outputs on these images are far, i.e.~$\loss (C(G(z)),C(G(z')))$ is large, where $\loss$ is some loss function such as cross-entropy. We show how to use our overpowered attack to fully circumvent DefenseGAN under the settings of ~\citet{athalye2018obfuscated} and~\citet{samangouei2018defensegan}.\footnote{There are a few key differences in hyperparameters for DefenseGAN used by~\citet{athalye2018obfuscated} and the original DefenseGAN paper~\citep{samangouei2018defensegan}. We describe the differences in the experimental section.}
This resolves an open challenge\footnote{
At the ICML 2018 best paper award talk, N. Carlini poses this as a challenging problem, see Min. 15 of talk.
\url{https://nicholas.carlini.com/talks/2018_icml_obfuscatedgradients.mp4}}
by~\citet{athalye2018obfuscated}.

We also show how our overpowered attack 
can be used to strengthen adversarial training. 
We set up a min-max problem in which the min player chooses the parameters of the classifier and the max player runs the overpowered attack against the classifier chosen by the min player. We call our proposed defense the \textit{Robust Manifold Defense}. 
If the range of the spanner is a good approximation to 
the true dataset, a successful defense against adversarial examples in the range of the spanner is sufficient to imply the lack of adversarial examples close to images in the dataset (by the triangle inequality).

Our robust optimization perspective towards adversarial training a la~\citet{madry2017towards}, 
illustrates the benefit of our proposed method: We are tremendously reducing the dimensionality of the space
that the max player is searching over. 
Rather than searching for adversarial examples inside balls around real images $x\in \mathbb{R}^n$, the max player is searching over pairs of ``latent codes'' $z,z' \in \mathbb{R}^k$. Lowering the dimensionality of the max player's search space renders the min-max optimization problem more benign. In particular, targeting this lower-dimensional space makes the max player more effective in identifying images in the range of the spanner on which the classifier's output differs, and in turn this renders adversarial training more effective. In practice, we may combine normal adversarial training  with adversarial training using our  overpowered attack, alternating between the two for multiple iterations to reap the benefits of both methods. We use our Robust Manifold Defense to obtain the most robust known MNIST classifier, significantly improving the state of the art against PGD attacks.

\subsection{Contributions}
Our contributions are summarized as follows:

\begin{itemize}
    \item  We propose a new type of attack that we call the \textit{overpowered attack}, which searches over pairs of adversarial images in the range of a spanner.
    
    \item  We show how to use our overpowered attack to {\em fully} circumvent DefenseGAN~\citep{samangouei2018defensegan}. 
This resolves an open challenge posed by~\citet{athalye2018obfuscated}. 
\item We show that our overpowered attack can be combined with adversarial training to increase the adversarial robustness of MNIST classifiers against white box attacks with bounded $\ell_2$ norm$=1.5$ from a state-of-the-art adversarial accuracy of $91.88\%$ using the TRADES algorithm~\citep{zhang2019theoretically} to an adversarial accuracy of $96.26\%$. This is a significant increase in a metric that has been proven challenging to improve previously.

    \item Finally, note that our overpowered attack creates new pairs of adversarial images (that are in the range of the spanner). We emphasize, however, that for evaluating both the circumvention of DefenseGAN and the robustness of our new MNIST classifier, we report adversarial robustness computed on the \textit{original test set images}. Our results are therefore directly comparable to all previous results in the literature.  
\end{itemize}

\section{Related work} 

There is  a deluge of recent work on adversarial attacks and defenses. Common defense approaches involve modifying the training dataset such that the classifier is made more robust~\citep{gu2014towards},~\citep{shaham2015understanding}, modifying the network architecture to increase robustness~\citep{cisse2017parseval}
or performing defensive distillation~\citep{papernot2016distillation}.
The idea of adversarial training~\citep{goodfellow2014explaining}
and its connection to robust optimization~\citep{shaham2015understanding,madry2017towards,sinha2017certifiable} leads to a fruitful line of defenses. 
On the attacker side,~\citep{carlini2017adversarial},~\citep{carlini2017towards} show different ways of overcoming many of the existing defense strategies. 

Our approach to defending against adversarial examples leverages GANs and VAEs~\citep{goodfellow2014generative,kingma2013auto} as spanners. GAT-Trainer by~\citep{lee2017generative} uses GANs to perform adversarial training but in a very different way from our work and without projecting on the range of a GAN. MagNet~\citep{meng2017magnet} and APE-GAN~\citep{APEGAN} have the similar idea of denoising adversarial noise using a generative model but use differentiable projection methods that have been successfully attacked by~\citep{carlini2017magnet}. Additionally, work by~\citep{song2018constructing} shows that generative models can be used to construct adversarial examples for classifiers, such that the classifier prediction is very different from a human prediction of the same image.


The most closely related work is DefenseGAN~\citep{samangouei2018defensegan}, which we have already compared our work to in length. One of our main contributions is to circumvent it using our overpowered attack.
The second related paper is PixelDefend~\citep{song2017pixeldefend}. This is similar to DefenseGAN except it uses PixelCNN generators. We note that \citet{athalye2018obfuscated} has already showed that PixelDefend can be circumvented.

\section{An (Ineffective) First Take at Adversarial Defense with Spanners} \label{sec:simple GAN defense}

Given a classifier $\class$ parameterized by a vector of parameters $\theta$, an initial idea might be to defend it by finding the projection of the input to be classified to the range of a spanner $G$, such as a Generator from a GAN or a Decoder from a VAE, and apply the classifier to the projection.
More precisely, for some hyper-parameter $\eta$ and given an input $x$, we perform the following procedure that we call \textit{Invert and Classify (INC)}:\footnote{Note that this is not a defense we propose to use, but a prop to develop our  attack in the next section.}
\begin{enumerate}
    \item Perform gradient descent in $z$ (latent code) space to minimize  $\norm{\gen(z)}{x}$. Let $z^*$ be the point returned by gradient descent. (Ideally, we would want $z^* \equiv \argmin_z\norm{\gen(z)}{x}$, but we settle with whatever gradient descent returns.)
    \item If the ``projection'' $G(z^*)$ of $x$ in the range of the spanner $G$  computed in Step 1 is far from $x$, i.e.~if $\norm{G(z^*)}{x} \geq \eta$,  we reject the input $x$ as ``unnatural,'' since it lies far from the range of the spanner.
    \item Otherwise, we apply our classifier on the projected input, outputting a class label according to   $\class(\gen(z^*))$.
\end{enumerate}

At first glance, this may seem like a reasonable defense, since first-order methods such as PGD or FGSM will be unable to compute the gradient through the projection step due to its non-differentiability. However, as we show in the following section, this is not the case. Indeed, the INC defense belongs to precisely the same class of techniques used by DefenseGAN~\citep{samangouei2018defensegan} and PixelDefend~\citep{song2017pixeldefend}, two defenses which were both found to be circumventable with a simple white-box adversary~\citep{athalye2018obfuscated}, albeit DefenseGAN had not been fully circumvented prior to our work.

\section{The Overpowered Attack}
\label{sec:attack on simple defense}

In the previous section, we considered the simplest way of exploiting neural network-based spanners to defend against adversarial attacks. We find, however, that this ``projection-based'' approach to defend classifiers is ineffective against white-box adversaries. In particular, we propose a new attack which we refer to as ``overpowered attack." We show that the overpowered attack successfully reduces the accuracy of the INC defense from the previous section to 0\%. Then, we show that we can slightly modify the overpowered attack to {\em fully circumvent} the DefenseGAN defense. 



\paragraph{Deriving the Attack.} Given some perturbation distance $\varepsilon$, one way to attack the $(\eta,G)$-INC defense of some classifier $C_\theta$ is to find a real image $x$ that is $\eta$ close to the range of $G$, as well as another input $x'$ that is both $\varepsilon$-close to $x$ and also within $\eta$ from the range of the spanner, so that the classifications of the projections $G(z),G(z')$ of $x,x'$ according to $C_\theta$ are significantly different. 

%
%

If such $(x,x')$ exist, however, then (by triangle inequality) there must exist $z$ and $z'$ such that $G(z)$ and $G(z')$ are $(2\eta+\varepsilon)$-close, yet $\class(\gen(z))$ and $\class(\gen(z'))$ are far. The following optimization problem captures the furthest $\class(\gen(z))$ and $\class(\gen(z'))$ can be under some loss function $\loss$ of interest (e.g.~cross-entropy), subject to our derived bound, in terms of $\varepsilon$ and $\eta$, on the distance between $G(z)$ and $G(z')$.
%
\begin{align}
    \sup_{z,z'} &~~\loss\left({\class(\gen(z))},{\class(\gen(z'))}\right),\\
    \text{s.t. }& \norm{\gen(z')}{\gen(z)}^2 \le (2\eta+\varepsilon)^2. \label{eq:constraint}
\end{align}
As per our discussion, the above optimization problem upper-bounds how much an attack to INC can change the  output. It is an upper bound in a very strong sense. In particular,
\begin{itemize}
    \item if the value of the above optimization problem is $V^*$ then this means that for {\em any} real image $x$, {\em any} $z$ returned by Step 1 of INC on $x$, {\em any} image $x'$ that is $\varepsilon$-close to $x$, and {\em any} $z'$ that is returned by Step 1 of INC on $x'$, the loss from the output of INC on $x$ and $x'$ is at most ${V^*}$;
    \item this means, in particular, that if $V^*$ is small, then INC suffers loss at most $V^*$ {\em even accounting for the fact that INC uses gradient descent in Step 1 to project to the range of $G$, and as such this step may be suboptimal};
    \item as such, the above optimization problem captures the worst loss that INC may suffer from an ``overpowered adversary,'' who can adversarially control also what happens in Step 1 of INC for different inputs.
\end{itemize} 
Summarizing the above points, the above optimization problem serves as an upper bound on the loss from both adversarial attacks and the suboptimality in Step 1 of INC.


%

Using our overpowered attack, we achieve the following.

\paragraph{Circumventing DefenseGAN.}
     DefenseGAN~\citep{samangouei2018defensegan} is a proposed adversarial defense technique that operates quite similarly to INC. \citet{athalye2018obfuscated} show that by using an attack known as Backwards-Pass Differentiable Approximation (BPDA), it is possible to reduce the adversarial accuracy of DefenseGAN from the claimed 96\% accuracy to 55\% accuracy. This attack was completed under an $\ell_2$ threat model with an $\ell_2$ per-pixel distortion budget of $0.005$. 
     
    
    By combining the overpowered attack that we outlined above with the expectation-over-transformation attack of~\citet{athalye2017synthesizing}, we can successfully reduce the classification accuracy of DefenseGAN to \textbf{3\%}, under the same perturbation budget and hyperparameter settings\footnote{In fact, we use the {\em average} $\ell_2$ perturbation budget reported by Athalye et al. as the {\em maximum}  one in our attack.}. We also evaluate our attack against DefenseGAN using the hyperparameter settings recommended in the original DefenseGAN paper~\citep{samangouei2018defensegan} and the perturbation budget in~\citet{athalye2018obfuscated} and find that the classification accuracy drops to \textbf{5\%}. In addition to resolving a major open problem, this attack demonstrates the potential of the latent-space attack as a more powerful attack for adversarial training. Please see Appendix~\ref{sec: hyperparams defensegan} for details about the attack hyperparameter settings.
    
     We stress the following. As outlined above, the overpowered attack identifies latent codes $z,z'$ whose images $G(z), G(z')$ are classified by $C_\theta$ as differently as possible. There is no guarantee that any of $G(z)$ or $G(z')$ is a real image or close to one. To circumvent DefenseGAN, we need instead to identify adversarial images that are $\varepsilon$-close to {\em real} images. To use the overpowered attack for this purpose, we need to force $G(z)$ and $G(z')$ to be appropriately close real images. We outline how this can be done in Section~\ref{sec:experiments}. In particular, our claimed reduction of DefenseGAN accuracy to \textbf{3\%} is measured in the standard way, i.e.~performing perturbations around {\em real images}, with the exact same perturbation budget under which~\citet{athalye2018obfuscated} reduce the DefenseGAN accuracy to 55\%.

\section{The Robust Manifold Defense}
\label{sec:minimax} \label{sec:less simple GAN defense}


The attack of the previous section should cause us to reconsider the use of INC and similar ``projection-based'' approaches for building adversarial defenses. At the same time, the effectiveness of the overpowered attack to resolve the outstanding challenge of circumventing DefenseGAN motivates us to use the overpowered attack within a robust optimization framework a la~\citet{madry2017towards} to build adversarial defenses. This is what we explore in this section, proposing for this purpose the following min-max formulation. The outer (inf) player of this formulation is setting the parameters of a classifier, while the inner (sup) player is searching for overpowered attacks:
\begin{align} \label{eq:constraint1}
    \inf_\theta\ \  &\mu \left(\sup_{\begin{minipage}[h]{2.8cm} \tiny $z,z'$:$\norm{\gen(z)}{\gen(z')}^2\le$\\\text{ }~~~~~~~~~~~~~~~~~~~~~~~~~$ (2\eta + 2\varepsilon)^2$\end{minipage}} \loss\left({\class(\gen(z))},{\class(\gen(z'))}\right) \right) +  (1-\mu) \left( \frac{1}{N}\sum_{i=1}^N \loss\left(y^{(i)},\class(x^{(i)})\right)\right). 
\end{align}
The mixing weight $\mu \in (0,1)$ is some hyperparameter that mixes between two objectives. The first measures how much an overpowered attacker can change the output of the classifier. The second  measures how well the classifier performs on the training set $\{x^{(i)}, y^{(i)}\}_{i=1}^N$. $\loss$ is the loss function that we are interested in  minimizing, e.g.~cross-entropy loss. 
Finally,  we choose $\varepsilon$ to be the allowed perturbation, and we choose $\eta$ according to the quality of our spanner $G$; in particular, we would like to choose $\eta$ so that real images are within $\eta$ from the range of the spanner. The quantity $(2\eta + 2\varepsilon)$ appearing in the constraint of the sup player has a similar justification as the justification of $(2\eta+\varepsilon)$ in our description of the overpowered attack in the previous section, namely: the existence of a true adversarial pair, consisting of a {\em real image} $x$ and a perturbed image $x'$ for which the classifier has high loss $\loss(C_\theta(x),C_\theta(x'))$ implies, by triangle inequality, the existence of a pair of latent codes $z$ and $z'$ such that $\norm{G(z)}{G(z')}\le 2\eta+2\varepsilon$ such that  $\loss(C_\theta(G(z)),C_\theta(G(z')))$ is high. Thus, if we want to protect against the existence of the former types of pairs $(x,x')$, it suffices to protect against the latter types of pairs $(z,z')$.


Our  adversarial defense approach has the following advantages/disadvantages in comparison to previous min-max formulations of adversarial training, such as~\citep{madry2017towards}, which search for adversarial examples around real images:

\begin{enumerate}
\item Notice that the sup player in our formulation searches for pairs $z,z'$ in {\em latent space}, which is typically much lower-dimensional compared to image space. As such, the sup player in our formulation faces an easier optimization problem than that facing the sup player in the standard min-max formulation of adversarial training, where the sup player is searching in image space. Given the challenging nature of min-max optimization, a decrease in the dimensionality of one of the two players of the optimization problem could provide big gains in our ability to get good solutions.


\item  As discussed in length in the previous section, the overpowered attack used by the sup player of our formulation is significantly more powerful than perturbing images from the training set. In effect, it allows the attacker to find adversarial {\em pairs of images} $x,x'$, such that neither $x$ nor $x'$ need to be in the training set, and which are close to each other, yet result in classifier outputs that are far. The increased power of the adversary could make a big difference in the robustness of the classifier, since, as also suggested by~\citet{madry2017towards}, endowing the sup player of adversarial training with the ability to perform stronger attacks could yield a stronger defense. Note also that it is meaningless to search over pairs of inputs $x,x'$ in the ambient space of images, as these are mostly garbage. Searching over pairs of images is only made possible by using the spanner. 

\item The disadvantage of our approach is that it needs a good spanner $G$. The better the spanner, the lower we could choose $\eta$ and the closer the range of $G$ would be to real images. The worse the spanner, the higher we would need to choose $\eta$ and the further the range of $G$ might be from real images. In this case, our attacker will be overly powerful, which could make our classifier overly defensive, which might decrease its adversarial robustness with respect to real images.

\end{enumerate}

We use our adversarial training procedure to train robust classifiers for the MNIST~\citep{lecun1998gradient} and CELEBA~\citep{liu2015faceattributes} datasets, and report our findings in Section~\ref{sec:experiments}. A main contribution of our approach is that we improve the adversarial robustness of MNIST classifiers against white box attacks with bounded $\ell_2$ norm$=1.5$ from a state-of-the-art adversarial accuracy of $91.88\%$ using the TRADES algorithm~\citep{zhang2019theoretically} to an adversarial accuracy of $96.26\%$.

\section{Experiments}\label{sec:experiments}

\subsection{Breaking DefenseGAN Additional Details}

In this section we show that we can use our latent space attack combined with the expectation-over-transformation attack~\citep{athalye2017synthesizing} to break DefenseGAN~\citep{samangouei2018defensegan}. It is worth noting that this attack 
has some modifications from the straightforward  overpowered attack. In the overpowered attack, we search for a pair $(z,z')$ such that $G(z), G(z')$ are close but the classifier makes different predictions on them. However, when attacking DefenseGAN, we do not have the freedom of choosing a pair of images which are classified differently. Instead we are given a \emph{fixed image} $x$ from a dataset and must adversarially perturb it such that DefenseGAN is fooled by the attack.

Our approach to breaking DefenseGAN is as follows: say we are given an image $x$ with true class $\ell$, and the GAN $G$. We wish to find a latent $\hz$ satisfying $\frac{\|G(\hz) - x\|_2}{784}\leq\deps$ (this corresponds to a an $\ell_2$ perturbation of 0.0051 per pixel, as in~\citet{athalye2017synthesizing}), such that when $G(\hz)$ is provided as input to the DefenseGAN mechanism, the classifier has different predictions for $x$ and $G(\hz)$.

We will now describe our algorithm for computing $\hz$. Assuming we are given the GAN $G$, and the classifier $C$, we solve the following optimization problem:
\begin{align}\label{eqn: latent space attack}
    \hz = \argmax_{z\in\R^k} \E_{\tau\sim\N(0,1)} &\left[\loss \left(C\left(G(z)+\tfrac{\sigma\tau}{\|\tau\|_2}\right) , \ell\right)\right],\\
    & \text{s.t.}\quad \dfrac{\|G(z) - x\|_2}{784}\leq \deps,
\end{align}
where $\loss\left(C(x'),\ell\right):= \argmax_{\ell'\neq \ell} C_{\ell'}(x') - C_{\ell}(x'),$ is the Carlini-Wagner loss~\citep{carlini2017towards} and $C_{\ell'}(x'),C_{\ell}(x')$ are the classifier logits for classes $\ell', \ell$ on the image $x'$. The random normal noise $\tau\sim\N(0,1)$ and the corresponding expectation over $\tau$ forms the expectation-over-transformation~\citep{athalye2017synthesizing} part of our attack. It is necessary to make our attack robust to noise implicit in the DefenseGAN procedure- if we provide as input an image $G(\hz),$ DefenseGAN may recover a latent code which is close to, but not exactly $\hz$. To make sure that the code recovered by DefenseGAN produces a misclassified image, we introduce noise of our own and average over it, which ensures that it is robust to noise in optimization introduced by DefenseGAN. To this end, we introduce the scaling constant $\sigma$ in~\eqref{eqn: latent space attack} which can be tuned to account for noise introduced by the DefenseGAN procedure.

By introducing a Lagrange multiplier $\lambda$, we can rewrite the above problem as an equivalent max-min optimization problem:
\begin{align}\label{eqn: maxmin defensegan}
    \max_{z\in\R^k}\min_{\lambda\geq 0} \E_{\tau} \left[\loss \left(C\left(G(z)+\tfrac{\sigma\tau}{\|\tau\|_2}\right) , \ell\right)\right] + \lambda \left( \deps - \frac{\|G(z) - x\|}{784}\right).
\end{align}

We find that this max-min optimization problem is tractable by doing gradient descent ascent in the primal variable $z$ and Lagrange multiplier $\lambda$. See Algorithm~\ref{alg:defense-gan} for pseudo-code detailing our attack and section~\ref{sec: hyperparams defensegan} in the Appendix for hyperparameters used in optimization.

\begin{figure*}
    \centering
    \includegraphics[width=0.4\columnwidth]{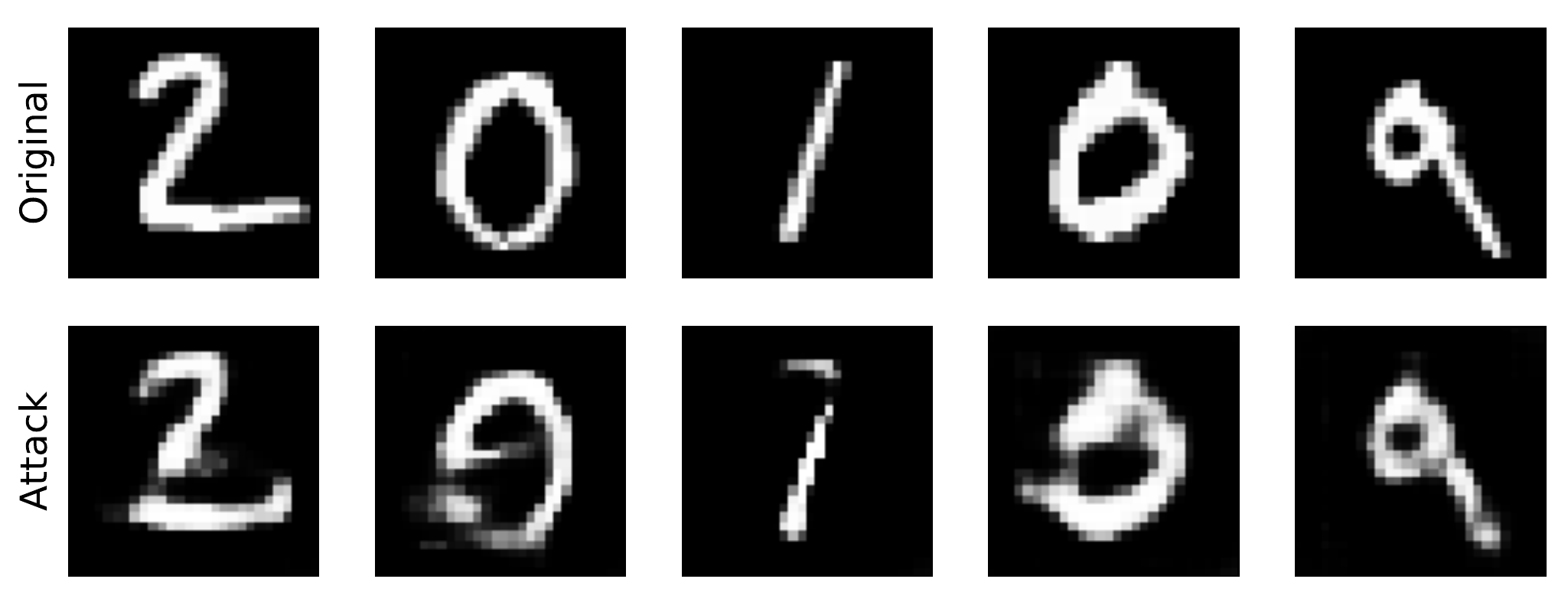}
    \caption{Attacks against DefenseGAN. The top row shows images from the MNIST test set and the bottom row shows their corresponding perturbed versions. The attacks constructed in the bottom row satsify the perturbation constraint imposed by~\citet{athalye2018obfuscated} and also successfully break DefenseGAN, i.e. are misclassified.}
    \label{fig:defensegan_attack}
\end{figure*}

\begin{algorithm}[t]
\caption{Pseudo-code for Defense-GAN break}
\label{alg:defense-gan}
\begin{algorithmic}
\STATE {\bfseries Input:} Image $x$, label $\ell$, classifier $C$, GAN $G$
\STATE {\bfseries Hyperparameters:} Step-size $\eta$, number of restarts $R$, number of gradient steps $T$, noise variance $\sigma$.\\
\STATE Find inverse of $x$, i.e., $z^* = \argmin_z \|x-G(z)\|_2.$
\STATE $\mathcal{L}(C(x'), \ell) := \argmax_{\ell'\neq \ell} C_{\ell'}(x') - C_{\ell}(x'),$ where $C_{\ell'}(x')$ is the logit for class $\ell'$ on the image $x'$.
\vspace{2pt}
\STATE // Do $R$ restarts and pick the best restart.
\FOR{$r\leftarrow 1$ to $R$ } 
    \STATE Set $z_r(0)\leftarrow z^*.$
    \STATE Set $\lambda_r(0) \leftarrow 10^3.$
    \vspace{2pt}
    \STATE // Do $T$ steps of gradient descent-ascent.
    \FOR{$t\leftarrow 0$ to $T-1$}
        \STATE Sample $\tau_j\sim\mathcal{N}(0,1), j=1,\cdots,100.$
        \vspace{2pt}
        \STATE loss$_r(t)\leftarrow \lambda_r(t) \cdot \left(0.0051-\frac{\|G(z_r(t)) - x\|}{784}\right) + \sum_{j=1}^{100} \mathcal{L}\left(C\left(G(z_r(t)) + \sigma\frac{\tau_j}{\|\tau_j\|}\right), \ell\right).$
        \vspace{2pt}
        \STATE $z_r(t+1)\leftarrow z_r(t) + \eta \nabla_z$ loss$_r(t)$. // ascent step in $z$.
        \vspace{2pt}
        \STATE $\lambda_r(t+1)\leftarrow \lambda_r(t) - \eta \nabla_\lambda$ loss$_r(t)$. // descent step in $\lambda$.
    \ENDFOR
\ENDFOR
\vspace{2pt}
\STATE $r^*\leftarrow \argmax_{r} \E_{\tau\sim\mathcal{N}(0,1)}\left[\mathcal{L}\left(C\left(G(z_r(T)) + \sigma\frac{\tau}{\|\tau\|}\right), \ell\right)\right]$ // restart that had largest CW-loss.
\STATE {\bfseries Output: $G(z_{r^*}(T))$.} 
\end{algorithmic}
\end{algorithm}
\paragraph{Empirical Results:}
When we evaluate our latent space attack~\eqref{eqn: latent space attack} with $\sigma = 0.5$ on the MNIST \emph{test set}, under the same hyperparameter settings for DefenseGAN used by~\citet{athalye2018obfuscated}, we find that DefenseGAN is robust to only \textbf{3\%} of our attacks. We emphasize that if $x_i$ is the $i^{th}$ image in the MNIST test set, then our attack $G(\hz_i)$ satisfies the perturbation constraint in the threat model. Of the 10000 images in the MNIST test set, we can find attacks such that 97\% of them fool DefenseGAN. Figure~\ref{fig:defensegan_attack} shows some examples of original images and perturbed images which break DefenseGAN.

\citet{samangouei2018defensegan} and \citet{athalye2018obfuscated}
have a few differences in hyperparameters employed within DefenseGAN- the number of gradient descent steps and number of restarts used to invert an image are two key parameters in DefenseGAN. \citet{samangouei2018defensegan} use 200 gradient steps with 10 random restarts, whereas~\citet{athalye2018obfuscated} use 20,000 steps. As noted in~\citep{samangouei2018defensegan}, increasing the number of gradient descent steps can decrease classification accuracy, since adversarial noise may be reconstructed. The use of only 200 gradient steps in DefenseGAN implies that the projection step has a lot of variance in the inversion step, i.e., the reconstructions obtained over multiple runs of DefenseGAN can vary a lot. To make our attack robust to this variance, we run a loop over different values of $\sigma=[1, 1.5, 2, 2.5, 3, 3.5]$ in~\eqref{eqn: latent space attack} and find that the classification accuracy of DefenseGAN drops to \textbf{5\%}. Please see Appendix~\ref{sec: hyperparams defensegan} for more details about the hyperparameters we used for DefenseGAN.

\subsection{Adversarial Training using the Overpowered Attack}
It is well known that a strong attack against a classifier can be used to improve its robustness through adversarial training~\citep{shaham2015understanding,madry2017towards}. Inspired by this, we verify that our proposed overpowered attack can be used to boost the robust accuracy obtained by Madry et al. on the MNIST dataset.

We first run the adversarial training procedure proposed by~\citet{madry2017towards} to get a base classifier $C^{(0)}$. The threat model allows white box access and perturbations whose $\ell_2$ norm is at most $\delta = 1.5$. During adversarial training, the adversary uses 40 step PGD with step size 0.075 and 1 random perturbation. Additionally, all pixel values are in the range $\left[0,1\right]$.

Starting with a robust classifier $C^{(0)}_\theta$ trained using the algorithm by~\citet{madry2017towards}, we employ our overpowered attack to improve its robustness in the following way:
\begin{enumerate}
    \item use the overpowered attack to find a batch of 50 pairs $\{(z_i,z'_i)\}_{i=1}^{50}$. We then modify the parameters of the network, $\theta$, by performing a single gradient descent step on the function 
    $10^{-2}\cdot \left(\frac{1}{50} \sum_{i=1}^{50}\loss(C^{(0)}_\theta(G(z_i)),C^{(0)}_\theta(G(z'_i)))\right),$
    where $\loss$ is the cross-entropy loss.
    \item Rerun the adversarial training algorithm in~\citet{madry2017towards} for 5 epochs, starting with the parameters obtained after Step 1.
    \item Repeat step 1 using $C^{(1)}_\theta$.
\end{enumerate}

We observe that running the above loop approximately 10 times returns a classifier whose robustness against white box access, bounded $\ell_2$ norm perturbations of the MNIST test set is 8\% more than the initial robust model $C^{(0)}$. We would also like to emphasize that the spanner model was trained only on images in the MNIST training set, and contains no information about the MNIST test set. Please see Section~\ref{sec:appendix_mnist} in the Appendix for more details about hyperparameters, train/validation set, etc.

\begin{figure*}
    \centering
    \includegraphics[width=0.4\columnwidth]{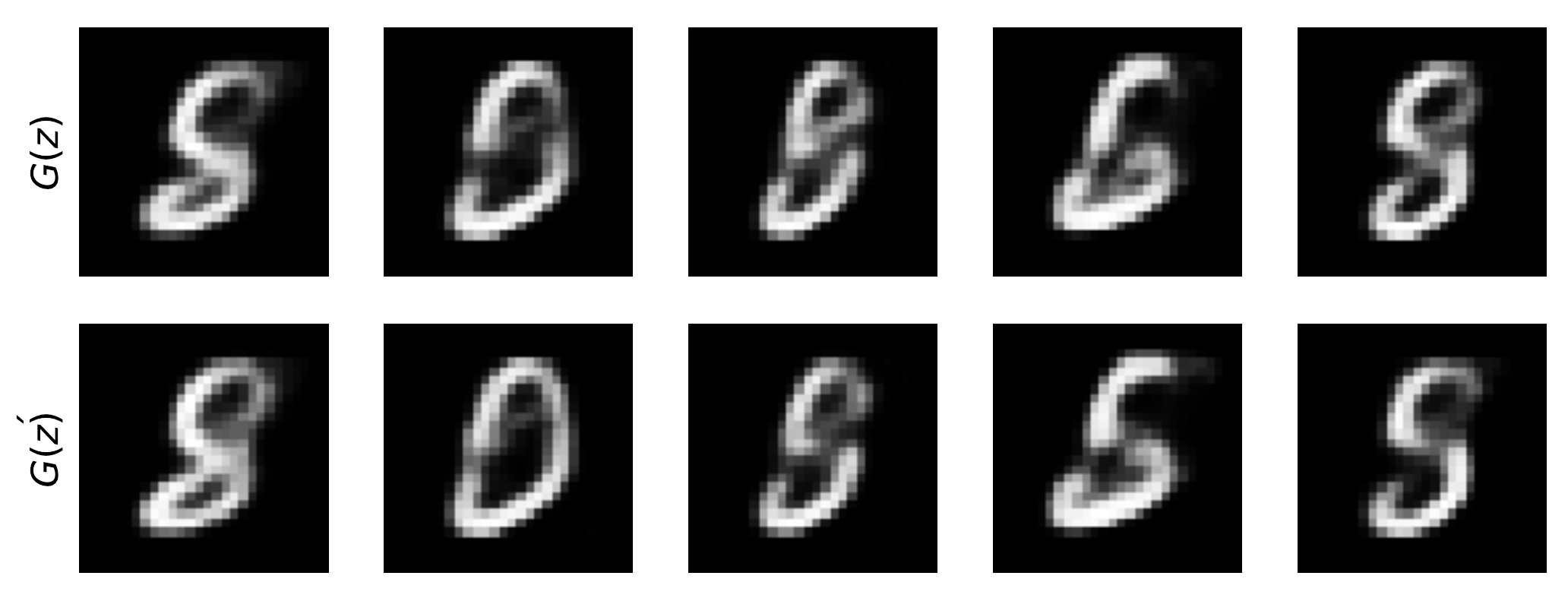}
    \caption{Pairs of images $G(z)$ and $G(z')$ generated by our overpowered attack, such that they are classified differently by a model trained using the adversarial training algorithm in~\citet{madry2017towards}. These images are used for adversarial training by penalizing the cross entropy loss of the classifier's prediction on pairs of images. We observe that using these images in conjunction with samples from~\citet{madry2017towards} improves robustness. Note that the final evaluation of robustness is performed on the actual MNIST test set images.}
    \label{fig:op_madry}
\end{figure*}

The intuition for why the overpowered attack helps robustness is as follows: the algorithm by~\citet{madry2017towards} finds adversarial perturbations on images in the training set. If it is run long enough, we observe that the model is capable of achieving 100\% robustness against the adversary on the training set, but the robustness on the test set plateaus at 90\% after roughly $10-20$ epochs. If we run the overpowered attack on this classifier, it will produce images as shown in Figure~\ref{fig:op_madry} that trick the classifier and are not based on images in the training set. Our hypothesis is that these samples allow us to meaningfully perturb the classifier, such that if we rerun the algorithm employed by Madry et al. starting from the perturbed classifier, we reach a new local minimum with improved robustness.

\begin{table}[]
        \centering
        \begin{tabular}{cccc}
        \hline
     Attack &  Madry et al. & TRADES & Ours\\\hline
     PGD (40 steps,$\delta=1.5$) & 89.93\% & 91.88\% & 96.26\%\\
     PGD (100 steps,$\delta=1.5$) & 89.87\% & 91.82\% & 96.25\% \\
    PGD (40 steps,$\delta=2.5$) & 76.09\% & 69.59\%  & 95.21\% \\
    PGD (100 steps,$\delta=2.5$) & 75.90\% & 68.86\%  & 95.16\% \\
    \hline
        \end{tabular}
        \caption{Robustness of different models against PGD adversaries with white box access for perturbations with $\ell_2$ norm $1.5, 2.5$ on the MNIST test set. Our algorithm improves upon the current state of the art by at least 5\%. Although it was trained against perturbations whose $\ell_2$ norm was at most $\delta=1.5$, we notice that the robustness does not fall significantly even when we increase the budget to $\delta=2.5$.}
        \label{tab: eval adv training}
    \end{table}
    
\paragraph{Results:} We report our empirical results in Table~\ref{tab: eval adv training}. We compare our model's robustness against models trained using the TRADES algorithm~\citep{zhang2019theoretically} with hyperparameter $\beta=6$ (we did a hyperparameter search over $\beta=1,6,10$) and the PGD training in~\citep{madry2017towards}. All models were trained against adversaries that had white box access and perturbations had bounded $\ell_2$ norm $=1.5$. The reported numbers are for bounded $\ell_2$ perturbations of images in the MNIST test set. It is interesting to see that although our model was trained using a threat model where the adversarial perturbation budget was $\delta=1.5$, it has good robustness even when the perturbation budget for the adversary on the test set is increased to $\delta=2.5$. 
    
\subsection{Adversarial Training on the CelebA Dataset}
\newcommand{\todo}{\textcolor{red}{\textbf{TODO}}}

\begin{figure*}
    \centering
    \includegraphics[width=0.7\textwidth]{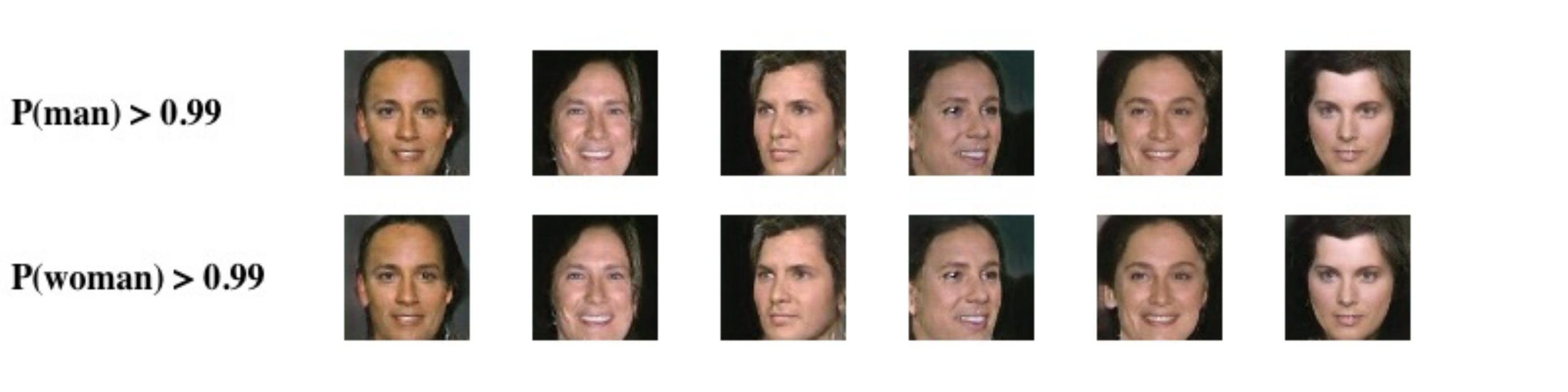}
    \caption{Pairs of images $G(z)$ and $G(z')$ generated by our overpowered attack on a non robust gender classifier on the CelebA dataset. 
    These images are very close but the confidence of the classifier changes drastically. These pairs of images are adversarial attacks for this classifier which lie on the manifold of a generator.}
    \label{fig:attack_simple_defense}
\end{figure*}

In this section we show that we can use the adversarial training procedure described in Section~\ref{sec:less simple GAN defense} to train a robust gender classifier on the CelebA dataset. 
Figure~\ref{fig:attack_simple_defense} shows randomly selected successful results of the overpowered attack against a non robust classifier protected by the INC algorithm. 

The attacks found with this optimization tend to yield images with semantically relevant features from both classes, and furthermore often introduce meaningful (though minute) differences between $G(z)$ and $G(z')$ (e.g. facial hair, eyes widening, etc.). Also, as confirmed by~\citep{athalye2018obfuscated}, none of these images actually induce different classifications on the end-to-end classifier, which we attribute to imperfections in the projection step of the defense (that is, since $G(z^*) \neq x$ exactly). However, we consider this implementation, and we robustify the classifier against this attack. We use a variant of the more complex min-max optimization proposed in Section~\ref{sec:less simple GAN defense}.
%


After 10,000 iterations, \textbf{100\%} of the images produced by the overpowered attack were valid, but with \textbf{22\%} of them inducing different classification, and an average KL divergence of \textbf{0.08}, showing that the classifier has softened its decision boundary. 



\begin{figure}  [htbp]
    \centering
    \includegraphics[width=0.75\columnwidth]{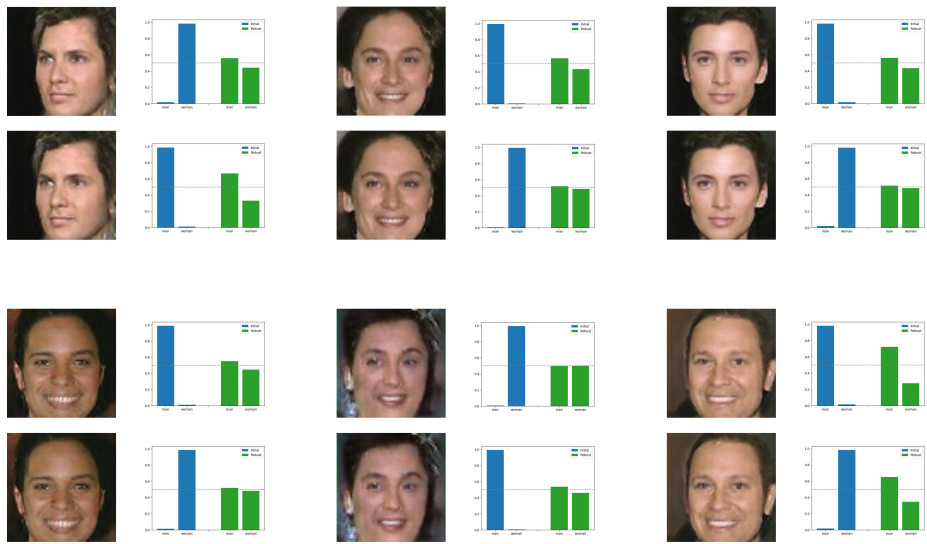}
    \caption{The softmax output of both the original (blue) and robust adversarially trained (green) classifier on the images generated by the attack on the non-robustified classifier. As shown, the robust projection defense makes the classifier reduce its confidence on such borderline images. }
    \label{fig:refeed_robust}
\end{figure}

We also feed the inputs generated by the overpowered attack on the initial classifier into the adversarially trained classifier. 
Figure~\ref{fig:refeed_robust} shows a randomly selected subset of these examples with their respective classifier output. Please see Section~\ref{sec:appendix celeba} in the Appendix for a more detailed description of the adversarial training procedure.

\section{Conclusion}
We show how we can use generative models
to create a new overpowered attack that searches over pairs of images in the spanner range. Our attack improves the state of the art for DefenseGAN resolving a challenging problem in the field. 

Further, we show that the generated attack images can be used to boost robustness of existing adversarial training procedures and can also be used to train robust classifiers that display natural uncertainty around decision boundaries. For MNIST we increase the $\ell_2$ robustness compared to the best previously known and for CelebA we show that there can be adversarial examples on the natural image spanner range. The main limitation of our approach is that it relies on a good spanner or generative model for the domain  of interest.



\clearpage
\bibliography{main}
\bibliographystyle{plainnat}

\appendix
\include{appendix}
\section{Hyperparameters for DefenseGAN break}\label{sec: hyperparams defensegan}
The generator and classifier architecture/checkpoint as well as an implementation of the attack can be found in the GitHub repository at \url{https://github.com/ajiljalal/manifold-defense}.

For the initial value of the variable $z$ in Eqn~\eqref{eqn: maxmin defensegan}, we first find the latent $z^*$ which produces an image in the range of $G$ which is closest to $x$, i.e., $z^*:=\argmin_{z\in\R^k}\|x-G(z)\|$ . We initialize $\lambda$ with the value $-1000$. To solve the optimization in Eqn~\ref{eqn: maxmin defensegan}, we use an Adam optimizer~\citep{kingma2014adam} with learning rate 0.05 for $z$ and an SGD optimizer with learning rate 10000 for $\lambda$.

\subsection{Hyperparameters for DefenseGAN}
We evaluate DefenseGAN under two hyperparameter settings:
\begin{itemize}
    \item \citet{athalye2018obfuscated}: In this setting, the projection operation in DefenseGAN uses 20,000 gradient steps with 20 random restarts. The reconstruction closest in $\ell_2$ to the original image is fed to the classifier as input. The projection step uses an SGD optimizer with learning rate=500 and momentum=0.7. Under this threat model, our combined attack (overpowered attack + expectation-over-transformation) reduces the accuracy of DefenseGAN to {\bf 3\%}. 
    \item \citet{samangouei2018defensegan}: In this setting, the projection operation in DefenseGAN uses 200 gradient steps with 10 random restarts. The reconstruction closest in $\ell_2$ to the original image is fed to the classifier as input. The projection step uses an SGD optimizer with learning rate=10 and momentum=0.7. Under this threat model, our combined attack of overpowered attack and expectation-over-transformation (EOT) reduces the accuracy of DefenseGAN to {\bf 5\%}. In this setting, we also grid-search over the EOT robustness parameter {\em for each image}.
\end{itemize}{}

\section{Hyperparameters for Boosting Madry's MNIST}\label{sec:appendix_mnist}

For the max-min optimization, we use the Yellowfin optimizer~\cite{zhang2017yellowfin} for $z$ and $\lambda$ with learning rate $10^{-4}$. We perform 500 updates for the overpowered attack. The classifier is trained with an adam optimizer with learning rate $10^{-4}$.

We use the first 55,000 images in the MNIST dataset for training the VAE and classifier. We use samples 55,000-60,000 as validation data. We use validation data to pick the classifer with maximum robust accuracy. The test set was the standard MNIST test set.

The MNIST classifier was the same used by Madry et al.~\cite{madry2017towards}, and we ran their code for 200 epochs against a PGD adversary, and the perturbations had bounded $\ell_2$ norm$=1.5$. The generative model was a VAE~\citep{kingma2013auto} from \url{https://github.com/pytorch/examples/blob/master/vae/main.py} except the decoder had architecture had 500 nodes in the first layer, 500 nodes in the second layer and the final output had width 784. We used a LeakyReLU as the non-linearity in the hidden layers.

The model definitions and checkpoints for the classifier and generator can be found at \url{https://github.com/ajiljalal/manifold-defense}.

The frequency with which we perform the overpowered attack is crucial to stabilize the adversarial training procedure. Using too many samples from the overpowered attack can cause the classifier to overfit to artifacts produced by the generator, and hence it may end up random guessing on the train/test set. We experimented with how many epochs of PGD training we should do before we perform the overpowered attack once. We tried running 3; 5; 8 epochs of regular PGD training in~\citet{madry2017towards} and then performing the overpowered attack to generate adversarial pairs. We found that running 5 epochs of PGD followed by one batch of overpowered attack samples provided best results. Selection of the best model was done by evaluating robustness on the validation set.
\section{Appendix for Adversarial Training on CelebA}\label{sec:appendix celeba}

In this section we show that we can use the adversarial training procedure described in Section~\ref{sec:less simple GAN defense} to train a robust gender classifier on the CelebA dataset. As described in section~\ref{sec:attack on simple defense}, we perform an overpowered attack on the standard invert-and-classify architecture.
We search for $z$ and $z'$ such that $G(z)$ and $G(z')$ are close but induce dramatically different classification labels. Recall that this involves solving a max-min optimization problem:
\begin{align*} 
\sup_{z,z'} \inf_{\lambda \le 0} \norm{\class(\gen(z))}{\class(\gen(z'))}^2 + \lambda \cdot (\norm{\gen(z)}{\gen(z')}^2 - (2\eta+\eps)^2).
\end{align*}
In practice, we set our $\ell_2$ constraint to $(2\eta+\eps)^2 \approx 2.46$, corresponding to an average squared difference of $2\cdot 10^{-4}$ per pixel-channel. We implement the optimization through alternating iterated gradient descent on both $\lambda$ and $(z, z')$, with a much more aggressive step size for the $\lambda$-player (since its payoff is linear in $\lambda$). The gradient descent procedure is run for 10,000 iterations. 
Because the $\ell_2$ constraint was imposed through a Lagrangian, we consider two $z, z'$ valid if the mean distance between the images is $< 5\cdot 10^{-4}$.
The optimization terminated with \textbf{93\%} of the images satisfying the $\ell_2$ constraint; within this set, the average KL-divergence between classifier outputs was \textbf{2.47}, with  \textbf{57\%} inducing different classifications. 
Figure~\ref{fig:attack_simple_defense} shows randomly selected successful results of the attack. 


The attacks found with this optimization tend to yield images with semantically relevant features from both classes, and furthermore often introduce meaningful (though minute) differences between $G(z)$ and $G(z')$ (e.g. facial hair, eyes widening, etc.). Also, as confirmed by~\citep{athalye2018obfuscated}, none of these images actually induce different classifications on the end-to-end classifier, which we attribute to imperfections in the projection step of the defense (that is, since $G(z^*) \neq x$ exactly). However, we consider this implementation, and we robustify the classifier against this attack. We use a variant of the more complex min-max optimization proposed in Section~\ref{sec:less simple GAN defense}:
\begin{align}
    \inf_\theta  \mu \left(\sup_{\begin{minipage}[h]{2.8cm} \tiny $z,z'$:$\norm{\gen(z)}{\gen(z')}^2\le$\\\text{ }~~~~~~~~~~~~~~~~~~~~~~~$(2\eta + 2\varepsilon)^2$\end{minipage}} \norm{\class(\gen(z))}{\class(\gen(z'))}^2\right)& + (1-\mu) \left( \frac{1}{N}\sum_{i=1}^N \loss\left(y^{(i)},\class(x^{(i)})\right)\right),
\end{align}
where $\loss$ is the cross-entropy loss.

\begin{figure}[ht]
    \centering
    \includegraphics[width=0.4\columnwidth]{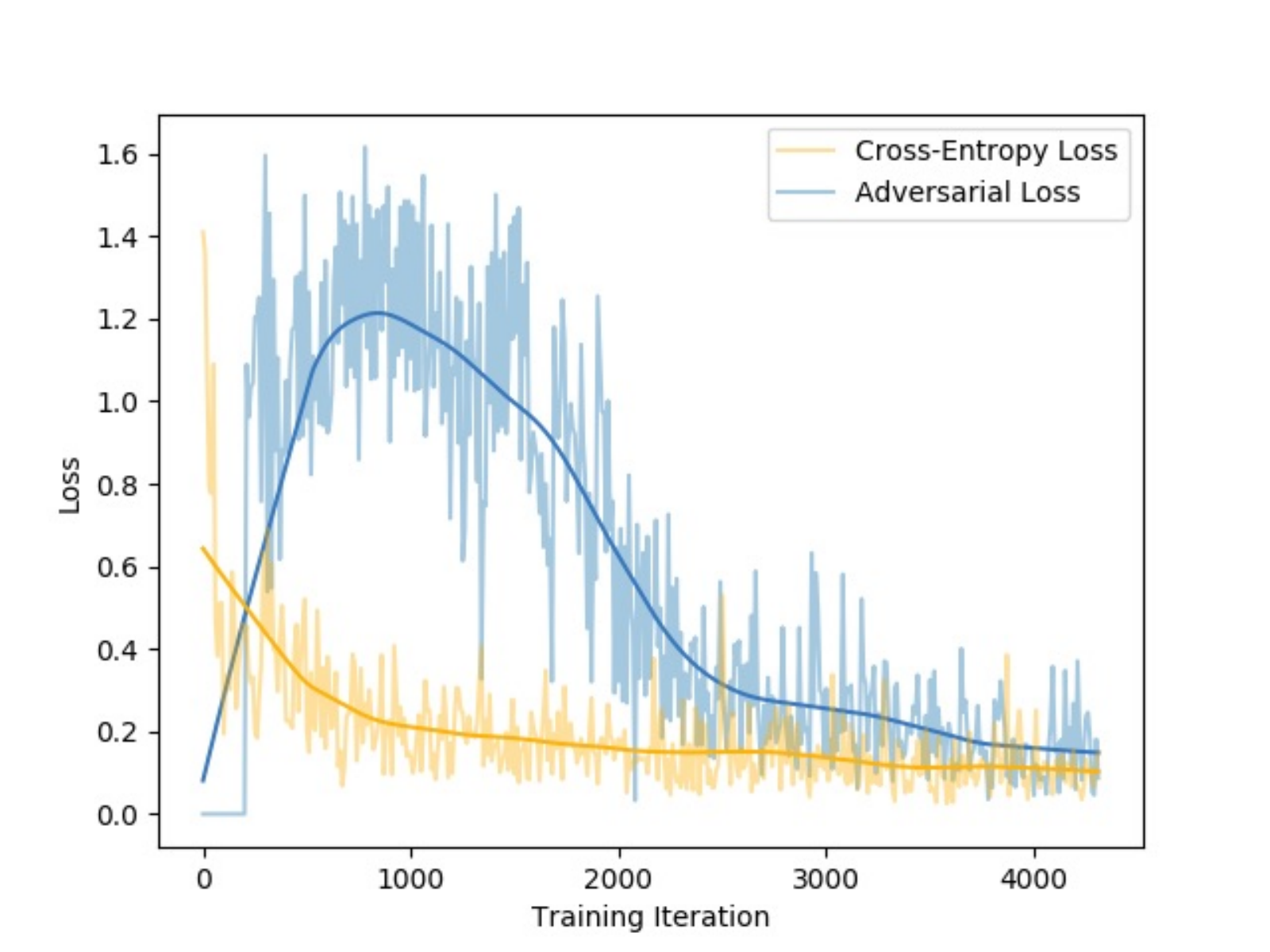}
    \includegraphics[width=0.4\columnwidth]{attack_progress}
    \caption{The cross-entropy and adversarial components of the loss decaying as training continues.}
    \label{fig:adversarial_loss_training}
\end{figure}

We implement this through \textit{adversarial training}~\citep{madry2017towards}; at each iteration, in addition to sampling a cross-entropy loss from images from the dataset, we also sample an adversariality loss, where we generate a batch of ``adversarial" inputs using 500 steps of the min-max attack, then add the final $\ell_2$ distance between the classification outputs to the cross-entropy loss. Please see Section~\ref{sec:appendix celeba} in the Appendix for more details about the adversarial training procedure. As shown in Figure~\ref{fig:adversarial_loss_training}, the classifier learns to minimize the adversary's ability to find examples. After robustifying the classifier using this adversarial training, we once again try the attack described earlier in this section for the same 10,000 iterations. Figure~\ref{fig:adversarial_loss_training} in the Appendix shows the convergence of the attack against both the initial and adversarially trained classifier for two values of $\eta^2$, showing the inefficacy of the attack on the adversarially trained classifier. After 10,000 iterations, \textbf{100\%} of the images were valid, but with \textbf{22\%} of them inducing different classification, and an average KL divergence of \textbf{0.08}, showing that the classifier has softened its decision boundary. 



In Table~\ref{table:robust_cla_acc} in we see that the robust classifier is effective against the overpowered latent space attack, which is an attack that is crafted for the \emph{INC protected classifier}. 

The adversarial training does not significantly impact classification accuracy over the standard classifier: on normal input data, the model achieves the same \textbf{97\%} accuracy undefended. We also feed the inputs generated by the min-max attack on the initial classifier into the adversarially trained classifier, and observe that the average classification divergence between examples drops to \textbf{0.007}, with only \textbf{18\%} of the valid images being classified inconsistently. 

Figure~\ref{fig:refeed_robust} shows a randomly selected subset of these examples with their respective classifier output.

We implement this through \textit{adversarial training}~\citep{madry2017towards}; at each iteration, in addition to sampling a cross-entropy loss from images from the dataset, we also sample an adversariality loss, where we generate a batch of ``adversarial" inputs using 500 steps of the min-max attack, then add the final $\ell_2$ distance between the classification outputs to the cross-entropy loss. As shown in Figure~\ref{fig:adversarial_loss_training} in the Appendix, the classifier learns to minimize the adversary's ability to find examples. After robustifying the classifier using this adversarial training, we once again try the attack described earlier in this section for the same 10,000 iterations. Figure~\ref{fig:adversarial_loss_training} in the Appendix shows the convergence of the attack against both the initial and adversarially trained classifier for two values of $\eta^2$, showing the inefficacy of the attack on the adversarially trained classifier. After 10,000 iterations, \textbf{100\%} of the images were valid, but with \textbf{22\%} of them inducing different classification, and an average KL divergence of \textbf{0.08}, showing that the classifier has softened its decision boundary. 
\begin{figure}[htbp]
    \centering
    \includegraphics[width=0.8\columnwidth]{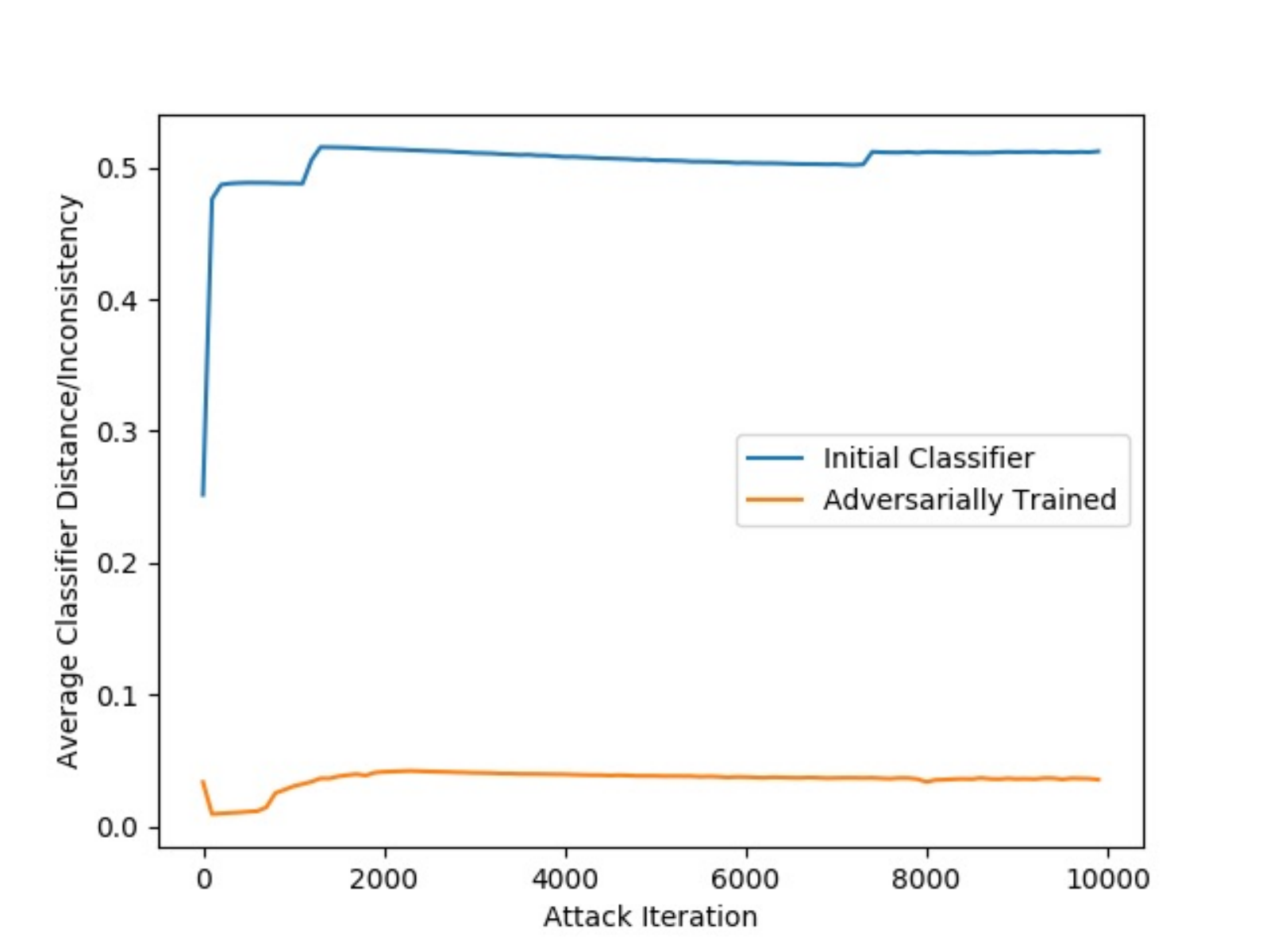}
    \caption{The average $||C(G(z)) - C(G(z'))||_2$ for pairs $(z, z')$ found by the attack.}
    \label{fig:attack_progress}
\end{figure}

In Table~\ref{table:robust_cla_acc}  we see that the robust classifier is effective against the overpowered latent space attack, which is an attack that is crafted for the \emph{INC protected classifier}. 

The adversarial training does not significantly impact classification accuracy over the standard classifier: on normal input data, the model achieves the same \textbf{97\%} accuracy undefended. We also feed the inputs generated by the min-max attack on the initial classifier into the adversarially trained classifier, and observe that the average classification divergence between examples drops to \textbf{0.007}, with only \textbf{18\%} of the valid images being classified inconsistently. 
\begin{table}[t]
\centering
  	\caption{Accuracy of the celebA classifier under different attacks: our overpowered attack, FGSM~\citep{goodfellow2014explaining}, BIM~\citep{kurakin2016adversarial}, PGD~\citep{madry2017towards} at various powers $\epsilon$. \textbf{NR} refers to \textbf{non robust classifier}, \textbf{NR+INC} refers to \textbf{non robust classifier with INC} and \textbf{R+INC} refers to \textbf{robust classifier with INC}.}
    \label{table:robust_cla_acc}
  \vskip 0.1em
  \begin{tabular}{@{}c|ccc@{}}
    \toprule
    \multirow{2}{*}{\textbf{Attack}} & \multicolumn{3}{c}{\textbf{CelebA}} \\
    & \textbf{NR}  & \textbf{NR+INC} & \textbf{R+INC} \\
    \midrule
    \midrule
    Clean Data              & 97\% & 84\% & 90\% \\
    Overpowered attack      & 0\% & 0\% & 90\% \\
    FGSM ($\epsilon = 0.05$)& 1\% & 82\% & 86\% \\
    FGSM ($\epsilon = 0.1$) & 0\% & 80\% & 77\% \\
    BIM ($\epsilon = 0.05$) & 0\% & 71\% & 85\% \\
    BIM ($\epsilon = 0.1$)  & 0\% & 63\% & 76 \% \\
    PGD ($\epsilon = 0.05$) & 0\% & 70\% & 84\% \\
    PGD ($\epsilon = 0.1$)  & 0\% & 62\% & 75 \% \\
    \bottomrule
    \bottomrule
  \end{tabular}
\end{table}
\subsection{Architecture}
\paragraph{Generator:} We use a BEGAN~\cite{berthelot2017began} and the Tensorflow repository from \url{https://github.com/carpedm20/BEGAN-tensorflow} as the generator.

\paragraph{Classifier} We modified the CIFAR10 classifier from \url{https://www.tensorflow.org/tutorials/images/deep_cnn} to work on the CelebA dataset~\cite{liu2015faceattributes} such that the final layer has 2 nodes for binary classification of gender.

\end{document}